\documentclass[journal]{IEEEtran}

\ifCLASSINFOpdf
\else
   \usepackage[dvips]{graphicx}
\fi
\usepackage{url}
\usepackage{graphicx}
\usepackage{subcaption}
\usepackage{amsmath}
\usepackage{amssymb}

\begin{document}

\title{EventF2S: Asynchronous and Sparse Spiking AER Framework using Neuromorphic-Friendly Algorithm}

\author{Lakshmi Annamalai, and Chetan Singh Thakur \IEEEmembership{Senior Member, IEEE}

}

\markboth{Journal of \LaTeX\ Class Files, Vol. 14, No. 8, August 2015}
{Shell \MakeLowercase{\textit{et al.}}: Bare Demo of IEEEtran.cls for IEEE Journals}
\maketitle

\begin{abstract}
Bio-inspired Address Event Representation (AER) sensors have attracted significant popularity owing to their low power consumption, high sparsity, and high temporal resolution. Spiking Neural Network (SNN) has become the inherent choice for AER data processing. However, the integration of the AER-SNN paradigm has not adequately explored asynchronous processing, neuromorphic compatibility, and sparse spiking, which are the key requirements of resource-constrained applications. To address this gap, we introduce a brain-inspired AER-SNN object recognition solution, which includes a data encoder integrated with a First-To-Spike recognition network. Being fascinated by the functionality of neurons in the visual cortex, we designed the solution to be asynchronous and compatible with neuromorphic hardware. Furthermore, we have adapted the principle of denoising and First-To-Spike coding to achieve optimal spike signaling, significantly reducing computation costs. Experimental evaluation has demonstrated that the proposed method incurs significantly less computation cost to achieve state-of-the-art competitive accuracy. Overall, the proposed solution offers an asynchronous and cost-effective AER recognition system that harnesses the full potential of AER sensors.
\end{abstract}

\IEEEpeerreviewmaketitle

\section{Introduction}

\IEEEPARstart{A}ddress Event Representation (AER) sensors \cite{12} \cite{13} \cite{14} \cite{25} \cite{26} \cite{27} \cite{28}, which aims to mimic biological eyes, has gained a lot of attention from different communities owing to its unique features. The inherent working principle of AER sensors results in the benefit of data sparsity, thus providing huge energy-saving. In contrast to a conventional camera, each pixel ($u,v$) of an event camera individually monitors the dynamics of the scene and records in the form of spiking asynchronous events as and when the change exceeds a pre-defined threshold. Events at pixel $(u,v)$ are generated as $4$-tuple $(t,u,v,p)$, where $t$ and $p$ represent the events' time and signed polarity of occurrence. 

The nature of AER data poses severe challenges in processing event data efficiently \cite{29} \cite{30} \cite{31} \cite{32} \cite{33}. The asynchronous event-driven paradigm of the AER sensor makes it a natural choice for Spiking Neural Networks (SNN) \cite{18} \cite{19} \cite{20} \cite{21} \cite{22}. SNN can be broadly classified into two approaches: Rate-coded SNN \cite{15} \cite{16} \cite{17} and temporal-encoded SNN \cite{1a} \cite{1b} \cite{1c} \cite{1d}. The fundamental problem of rate-coded SNN is that it results in tremendous spike signaling cost and latency. Unlike the above technique, temporal-encoded SNN in the AER domain involves hand-crafted spatiotemporal feature extraction followed by a spiking recognition layer. The feature extraction phase of these networks makes them challenging to implement in the temporal spiking domain. Recently a learnable temporal-coded feature extraction has been proposed in \cite{3a}. However, the major limitation of this method comes in the event encoding process, which incurs a memory cost of $O(N^2)$, thus making its implementation non-trivial. 

Motivated by the fact that neurons in the visual cortex respond in asynchronous mode with minimal energy cost, we asked ourselves whether a similar hypothesis could be achieved in AER recognition systems. Towards this, we introduce a novel single-spike object recognition system which we refer to here as EventF2S (Event Camera First-To-Spike Spiking Network). 

EventF2S consists of two main components: 1) Sparse Spiking Temporal Encoding (SS-TE) layer: This layer provides a neuromorphic implementable algorithm to encode raw events into as few informative early spikes as possible. The SS-TE encoding scheme delivers a drastically reduced number of spikes to computationally heavy downstream applications. 2) Spiking Recognition Network: It consists of a group of differentiable spiking neurons that adapts the First-To-Spike protocol for inter-neuron communication. The learnable spiking neural dynamics enable us to backpropagate the error deep down the network and achieve competitive accuracy with single-spike input encoding. By incorporating the First-To-Spike communication protocol, we aim to achieve further reduction in spike signaling.

To summarize, the significant contributions of EventF2S are as follows: i) \textbf{Asynchronous processing:} EventF2S exhibits asynchronous processing at the input and output layers. The system processes the events asynchronously as they arrive at the input layer. At the output layer, the winner neuron fires asynchronously once it has received sufficient spikes to activate it. Note that there is no pre-defined accumulation time; ii) \textbf{Sparse spiking:} EventF2S curtails the number of spikes to one per pixel and one per neuron at the encoding, feature extraction, and classification layers; iii) \textbf{Neuromorphic Friendly:} By designing the solution to be end-to-end neuromorphic hardware implementable, we enable EventF2S to inherit brain's efficiency and computational mechanisms.

\section{Related Work}

As the proposed framework is a temporal spiking neural network, we have not ventured into the survey of rate-coded spiking networks. Instead, we provide a survey specifically in the domain of temporal coding SNN, which falls under the following two categories: cortical mechanism-based \cite{1a} \cite{1b} \cite{1c} \cite{1d} \cite{1f} \cite{1g} \cite{1h} and statistical learning-based \cite{2a} \cite{37} \cite{38}. 

HMAX \cite{1a} is one of the most popular hierarchical models structured as simple (linear) and complex (non-linear) layers S1, C1, S2, and C2 with SVM as a classifier. Algorithm of \cite{1b} extracted position and size invariant features for human posture recognition. Zhao. et al. \cite{1c} proposed an event-driven neuron model in the S1 layer with a forgetting mechanism and replacing max-pooling with MAX operation at the C1 layer. S2 and C2 layers were discarded. The generated features converted to single-spike encoding have been classified using a Tempotron-based classification layer. Orchard. et al. \cite{1d} came up with an HMAX-based hierarchical spiking architecture. The winner-Take-All strategy was implemented in the C1 layer to realize Max operation. However, the learning layers used statistical methods, thus losing the spiking nature.

Liu. et al. proposed \cite{1f} proposed MuST (Multi-Scale spatiotemporal) feature representation based on Leaky Integrate and Fire (LIF) neurons followed by an unsupervised recognition approach. MuST features are converted into temporal coded spikes using a natural logarithm temporal coding scheme. Patterns are learned from these spatiotemporal spikes in an unsupervised fashion with Spike Timing Dependent Plasticity (STDP). The architecture of \cite{1g} consists of two modules: Hand-crafted multi-spike encoding followed by tempotron-based SNN learning. Though this method was fully temporally encoded, multiple spikes were required to achieve good accuracy as learning is implemented only in the classification stage.

\cite{1h} includes spatiotemporal feature extraction and a novel learning algorithm. The feature extraction phase is similar to that of \cite{1g}. As part of spike learning, they have introduced a novel learning algorithm known as the Segmented Probability Maximization (SPA) algorithm, which classifies based on the firing rate of neurons in the classification layer. Though \cite{1h} was able to achieve better accuracy than \cite{1g}, the rate coding of the decision layer discredits its use in the temporal domain.

Recently, \cite{3a} ventured into a temporal-encoded spiking object recognition system, which has three modules: pre-processing of events, temporal coding of events, and object recognition spiking network. The spike encoding involves the accumulation of event stream into an image frame, $D(x,y)$, over a period of time from $t_0$ to $t_k$, which incurs a memory cost of $O(N^2)$. As and when the value at pixel $(x,y)$ of $D(x,y)$ exceeds a pre-defined threshold, the particular input neuron spikes. This will incur additional memory costs, thus rendering it not suitable for neuromorphic implementation.

\section{Sparse Temporal-Encoded EventF2S Architecture}

The proposed spiking architecture is structured as follows: Single Spike Temporal Encoding (SS-TE) layer and a learnable First-To-Spike encoded object recognition spiking network. The SS-TE layer serves as the input layer of the spiking network. The recognition network has feature extraction convolution layers followed by a classification layer. 

\subsection{Single Spike Temporal Encoding (SS-TE) Layer}

SS-TE layer densely covers the pixels of the AER sensor, performing signal-selective single spike per pixel encoding. The new sensing modality of the AER sensor has made it more susceptible to noise. This results in an abundant number of events, potentially overloading the computation of downstream applications. The noisy events of the AER sensor broadly fall under two categories: Type I noise, which is uncorrelated with spatiotemporal neighbor events, and Type II noise, which exhibits self-temporal correlation.

To address this issue, we implement event encoding as a collection of parallel asynchronous continuous-time LIF (Leaky Integrate-and-Fire) filters whose spatial sensitiveness is defined to respond best to spatiotemporally correlated events while penalizing the Type II noise. The degree of penalization is inversely proportional to the frequency. 

\subsubsection{SS-TE Encoding Function}

The function of SS-TE encoding can be derived as furnished here. Events are modeled as a summation of impulse signals in continuous time (Eq. \ref{eq:1a}),

\begin{equation}
E(t)=\sum_{k=1}^{\inf}p_k\delta(t-t_k)\delta_{x_k}(x)
\label{eq:1a}
\end{equation}

Where $\delta(t)$ is the Dirac delta function, $p=\pm1$ is the polarity depending on the scene brightness change, $x=(u,v)$ is the discrete pixel address, $t_k$ is the timestamp of occurrence of the $k^{th}$ event, relative to the global clock and $\delta_{x_k}(x)$ is the Kronecker delta function which is $=1$ at $x=x_k$. 

Considering the events $E_{x_i^-,\vartriangle\tau}(t)$ that occurred on a local neighborhood $x_i^-=\{(u,v):|u-u_i|<\beta,|v-v_i|<\beta,x_i\nexists{x}$ centered at $i^{th}$ pixel and activated at $[t_i,t_i-\Delta\tau]$, it can be represented as a monodimensional signal in continuous time as follows,

\begin{equation}
E_{x_i^-,\vartriangle\tau_i}(t)=\int_{t_i-\Delta\tau}^{t_i}\delta(t-\tau)\delta_{x_i^-}(x)d\tau
\end{equation}

The events $E_{x_i,\vartriangle\tau}(t)$ that occured at the $i^{th}$ pixel in the time duration $[t_i,t_i-\Delta\tau]$ can be expressed as 

\begin{equation}
E_{x_i,\vartriangle\tau}(t)=\int_{t_i-\Delta\tau}^{t_i}\delta(t-\tau)\delta_{x_i}(x)d\tau
\end{equation}

Define the following continuous time single dimensional low pass filter $h(t)=e^{-\frac{t}{\tau_{c}}}$, with cut-off frequency determined by $\tau_{c}$. In order to remove the Type I and II noise, we need to maximize $h(t)\otimes{E_{x_i^-,\vartriangle\tau}}(t)$ and minimize $h(t)\otimes{E_{x_i,\vartriangle\tau}}(t)$ respectively. Summarizing, we need to estimate events such that the following objective function is maximized,

\begin{equation}
h(t)\otimes{E_{x_i^-,\vartriangle\tau}}(t)-h(t)\otimes{E_{x_i,\vartriangle\tau}}(t)
\end{equation}

The given objective function can be realized with a set of parallel LIF filters whose output is modeled as a convolution between asynchronously sampled continuous time filter $h(t)$ and the events $\left[E_{x_i^-,\vartriangle\tau}(t), E_{x_i,\vartriangle\tau}(t)\right]$ weighted with $\left[+1,-1\right]$. The signal spikes are estimated as a subset of AER spikes emitted by LIF neurons whose corresponding membrane voltage crosses a threshold. After firing, the particular neuron shunts all its incoming spikes till a classification decision is reached by the recognition spiking network. Thus, SS-TE encoding hugely reduces the number of spikes presented to the recognition spiking network.

\subsection{Learning by First-To-Spike Spiking Network}

In the learning stage, a single spike per pixel emitted by the LIF neurons of the SS-TE layer is transmitted to the convolution-based feature extraction layers arranged in a feed-forward fashion. The learned multi-channel feature spikes are fed to the classification layer, which consists of $C$ (number of classes) neurons. Once a neuron spikes, temporal inhibition is performed, which will force the active neuron to enter into a dormant stage. As a result, this promotes sparsity in spike representation.

\subsubsection{Neuron Dynamics}

This section describes the neural and synaptic dynamics of the neurons which form the building block of our spiking recognition network. Inspired by \cite{36}, we define our recognition network with a set of non-leaky IF neurons with synaptic kernel function $\Theta(t-t_i)e^{-\frac{(t-t_i)}{\tau_m}}$, where $\Theta(t-t_i)=1$ if $t\geq{t_i}$ and $0$ otherwise. 

As given in \cite{36}, the relation between input and output spike time when the neuron spikes are given as follows,

\begin{equation}
e^{t_{out}}=\frac{\sum_{i\in{C}}w_ie^{t_i}}{\sum_{i\in{C}}w_i-1}
\label{eq:1b}
\end{equation}

Where $C$ is the number of causal spikes which causes the neuron to spike and $w_i$ is the weight of the $i^{th}$ spike. The relation between the neuron's first spike time and input spike time given by Eq. \ref{eq:1b} is differentiable with respect to the weights, hence making the network trainable. As per Eq. \ref{eq:1b}, $\sum_i{w_i}$ should be greater than one for a neuron to spike. To ensure a sufficient number of neuron spikes, the following constraint has been imposed on weights as given in \cite{36}. 

\subsubsection{Loss Function}

The network is trained such that the output neuron belonging to the correct class fires early. To achieve the same, the following differentiable cross-entropy loss function is defined, which is minimized using the standard Gradient Descent technique.

\begin{equation}
-\sum_{k=1}^{C}ln\frac{e^{-z_k}}{\sum_i{e^{-z_i}}}
\label{eq:1c}
\end{equation}

Where $z_k$ is the output of $k_{th}$ output neuron, the negative of the output is used in Eq. \ref{eq:1c} in order to make the correct class neuron fire earlier.

\section{Experiments and Results}

In this section, we conduct two types of experiments: 1) To showcase the capability of the SS-TE encoding layer and 2) To bring out the recognition capability of the overall EventF2S framework. The neuromorphic datasets considered are \cite{9}, \cite{8}, N-MNIST \cite{1d-1} and MNIST-DVS \cite{39}.

\subsection{SS-TE Experiments}

This section highlights the benefits reaped by prefixing the recognition network with the SS-TE encoding layer. 

\subsubsection{Qualitative Analysis}

\begin{figure}
\centering
\includegraphics[width=0.48\linewidth,keepaspectratio=true]{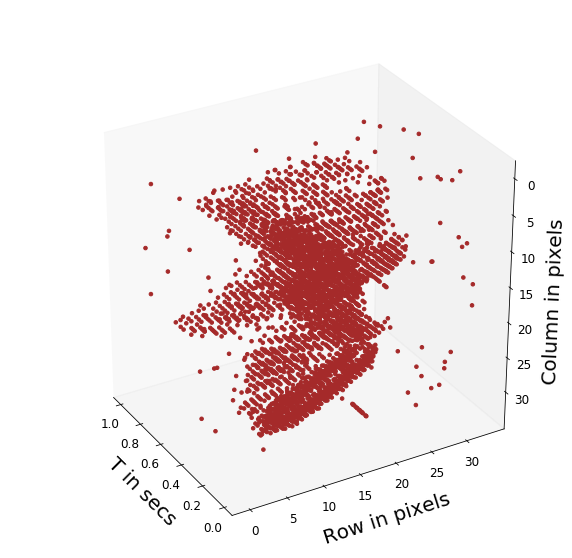}
\includegraphics[width=0.48\linewidth,keepaspectratio=true]{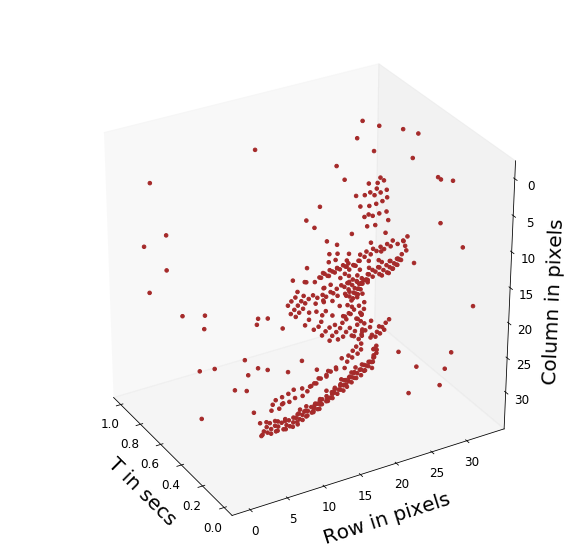}
\caption{Left: Raw event data of digit $5$ drawn from neuromorphic NMNIST dataset. Right: SS-TE encoding. SS-TE output being visually recognizable reinstates that SS-TE retains necessary information with just a single spike per pixel.}
\label{fig:SS-TE}
\end{figure}

This section provides qualitative evidence to showcase the effectiveness of the SS-TE encoding layer in extracting sparse and informative spatiotemporal events from raw events. Fig. \ref{fig:SS-TE} shows the 3D plot of the raw input event stream along with its SS-TE encoded output event stream. The plot demonstrates that despite the substantial decrease in the number of output events, the pattern of the digit $5$ is clearly visible in the SS-TE encoded output. The result highlights the capability of SS-TE to retain relevant information while discarding less informative events.

\begin{table*}[h!]
\centering
\begin{subfigure}{0.33\linewidth}
\centering
\begin{tabular}{ll|ll|ll}
\hline
\multicolumn{2}{c|}{\textbf{SS-TE}} & \multicolumn{2}{c|}{\textbf{Density} \cite{10}} & \multicolumn{2}{c}{\textbf{BAF} \cite{11}} \\ \hline
TP      & FP     & TP     & FP  & TP     & FP  \\ \hline
\textbf{0.98}   & \textbf{0.37}  & 0.99   & 0.99 & 0.99  & 0.99  \\
\textbf{0.95}   & \textbf{0.32}  & 0.89   & 0.85 & 0.99  & 0.99  \\
0.74   & 0.23   & 0.74   & 0.71 & 0.99  & 0.99  \\
0.39   & 0.14   & 0.60   & 0.55 & 0.99  & 0.99  \\
0.25   & 0.09   & 0.53   & 0.43 & 0.99  & 0.99  \\ \hline
\end{tabular}
\end{subfigure}
\begin{subfigure}{0.33\linewidth}
\centering
\begin{tabular}{ll|ll|ll}
\hline
\multicolumn{2}{c|}{\textbf{SS-TE}} & \multicolumn{2}{c|}{\textbf{Density} \cite{10}} & \multicolumn{2}{c}{\textbf{BAF} \cite{11}} \\ \hline
TP      & FP     & TP     & FP  & TP     & FP  \\ \hline
\textbf{0.99}   & \textbf{0.48}  & 0.99  & 0.99 & 1.119  & 0.844  \\
\textbf{0.92}   & \textbf{0.40}  & 0.87  & 0.85 & 1.119  & 0.844  \\
0.69   & 0.28  & 0.74  & 0.71 & 1.119  & 0.844  \\
0.27   & 0.15  & 0.61  & 0.85 & 1.119  & 0.844  \\
0.23   & 0.12  & 0.55  & 0.42 & 1.119  & 0.844  \\ \hline
\end{tabular}
\end{subfigure}
\begin{subfigure}{0.33\linewidth}
\centering
\begin{tabular}{ll|ll|ll}
\hline
\multicolumn{2}{c|}{\textbf{SS-TE}} & \multicolumn{2}{c|}{\textbf{Density} \cite{10}} & \multicolumn{2}{c}{\textbf{BAF} \cite{11}} \\ \hline
TP      & FP     & TP     & FP  & TP     & FP  \\ \hline
0.99   & 0.54  & 0.99  & 0.99 & 1.119  & 0.844  \\
\textbf{0.94}   & \textbf{0.46}  & 0.86  & 0.85 & 1.119  & 0.844  \\
0.68   & 0.30   & 0.72  & 0.70 & 1.119  & 0.844  \\
0.27   & 0.15   & 0.60  & 0.54 & 1.119  & 0.844  \\
0.23   & 0.13   & 0.54  & 0.42 & 1.119  & 0.844  \\ \hline
\end{tabular}
\end{subfigure}
\caption{Analysis of denoising methods in terms of True Positive (TP) and False Positive (FP). Left: $0.9$ SNR , Middle: $0.97$ SNR, Right: $0.98$ SNR . SS-TE encoding passes most of the signals (high TP) while eliminating all noisy events (low FP)}
\label{table:single}
\end{table*}

\subsubsection{Denoising Experiments}
Though the primary focus of our research is the object recognition spiking task, we include the performance of our SS-TE encoding layer on the denoising task as a means to validate that the proposed solution retains only the informative single spike for each pixel. Denoising was evaluated on a publicly available noise dataset, recorded by a Davis346 camera under bright and dark conditions \cite{8}. To evaluate the denoising performance, the noise events were mixed with pure event signals. A conventional camera video from \cite{8} was converted to pure events using the v2e algorithm \cite{9}.

Table. \ref{table:single} shows the comparison of the proposed encoding with state-of-the-art event denoising methods \cite{10} \cite{11} in terms of True Positive (TP) and False Positive (FP) rates at different SNR. The results demonstrate the effectiveness of SS-TE in achieving a high True Positive (TP) rate for a low False Positive (FP) rate across various Signal-to-Noise ratios (SNR). SNR is adjusted by varying the number of noisy events added. 

\subsubsection{Sparse Temporal Spiking}

The spikes generated and transmitted in a spiking network should be as low as possible to reap the full potential of the AER sensor. The average number of spikes per pixel ($N_s$) generated by the AER sensor, calculated across test samples, amounts to $49$ and $50$ for NMNIST and MNIST-DVS, respectively. SS-TE encoding layer reduces the spike count to just one per pixel as against $49$ and $50$.

\subsection{Recognition Experiments}

For evaluation of our primary object recognition task, the network topology used is as follows: Two convolution layers of $32$ and $16$ channels, each with a kernel size of $5\times{5}$ and stride of $2$, followed by a classifier layer of $10$ neurons. $10$ neurons represent the number of classes. During the training process, the learning rate was set to $1e^{-2}$ for the initial $50$ epochs, and later it was reduced to $1e^{-3}$ for the remaining $50$ epochs. 

\subsubsection{comparison in terms of Recognition Accuracy}

We have compared our approach with \cite{1b}, \cite{1c}, \cite{1d}, \cite{1d-1} \cite{1e} \cite{1f} \cite{2a} \cite{2b}. We have restricted our comparison to non-rate-coded spiking networks as the rate-coded networks provide good accuracy at par with ANN at the cost of huge energy consumptions as compared to temporal-coded networks. Results from the comparison are summarized in Table. \ref{table:accuracy} in terms of accuracy. It could be observed that EventF2S achieves good accuracy at par or above the state-of-the-art AER recognition networks. 

The performance of EventF2S in terms of accuracy is below \cite{3a}. While accuracy is acknowledged as an important metric, the specific goal of the proposed EventF2S is to provide a neuromorphic-friendly solution. This allows us to leverage the benefits of neuromorphic hardware, such as asynchronous processing and low cost. The encoding process mentioned in \cite{3a} incurs additional memory cost, which is a major drawback for hardware implementation. Additionally, the processing primitives of the encoding layer of \cite{3a} do not align with the device physics of neuromorphic hardware, further limiting the application of \cite{3a}. Thus, EventF2S is a promising asynchronous approach for low-cost edge applications.

\begin{table}[]
\centering
\begin{tabular}{c|cc|c|cc}
\hline
\textbf{Method}           & \textbf{NM} & \textbf{MD}  & \textbf{Method}  & \textbf{NM} & \textbf{MD}\\ \hline
\cite{1b} & - & 0.63 & \cite{1c} &  0.85 & 0.768            \\ 
\cite{1d} & 0.711 & 0.55 & \cite{1d-1} & 0.83 & - \\
\cite{1e} & 83.44 & - & \cite{1f} & 0.89 & 0.79 \\
\cite{1g}(ss) & 0.79 & -  & \cite{1g}(ms) & 0.93 & 0.77 \\ \hline
\cite{2a} & 0.704 & 0.746  & \cite{2b} & 0.80 & 0.80 (full) \\ \hline
EventF2S & \textbf{0.94} & \textbf{0.84}  & EventF2S$^{2s}$ & \textbf{0.97} & -  \\ \hline
\end{tabular}
\caption{Comparison of recognition accuracy with state-of-the-art methods on NMNIST (NM) and MNIST-DVS (MD) datasets. Accuracy of other methods are as reported in \cite{1f} \cite{2a} \cite{1h} \cite{1g} (MD: 100ms). The proposed method gives better accuracy than other non-rate-coded AER spiking networks. EventF2S$^{2s}$ refers to SS-TE emitting maximum two spikes per pixel. Accuracy increases with the number of spikes, which can be set as per the application requirement.}
\label{table:accuracy}
\end{table}

\subsubsection{Computation Efficiency}

In this section, we perform experiments to analyze the proposed EventtF2S in terms of computation cost in comparison with \cite{1c}, \cite{1d}, \cite{1f} \cite{1g} and \cite{1h}. Initial feature extraction layers contribute to the bulk of the computational complexity as they operate on larger spatial resolution. Hence, we evaluated the computation cost as the number of weight multiplications that takes place in the first layer of the feature extraction stage. 

To estimate computation cost, we multiply the number of events input into the first layer ($N_e$) with the number of computations incurred per event ($N_c$). Each event was assumed to be surrounded densely by neighboring events. The values of $N_e$ for the raw event and SS-TE encoding are $(4203,1156)$ and $(27320,16384)$ for NMNIST (NM) and MNIST-DVS (MD), respectively. $N_e$ is calculated as the mean of the number of input events that enter the feature extractor across the test samples. 

The results are provided in Table. \ref{table:compute} makes it evident that the computation incurred at the first layer of EventF2S is far lesser as compared to other methods. State-of-the-art methods perform feature extraction on every input event, thus increasing the computational complexity. By making the feature extraction learnable, we were able to reduce the number of computations drastically with single-spike encoding.

\begin{table}[]
\centering
\begin{tabular}{c|cc|c|cc}
\hline
\textbf{Methods} &    \textbf{NM}      &      \textbf{MD} & \textbf{Methods} &    \textbf{NM}      &      \textbf{MD}     \\ \hline
\cite{1c}, \cite{1h}               &      2.75           &           17.92  & \cite{1d}                 &    2.47             &         16.06    \\ 
\cite{1g}                 &        $N_a\times$2.47      &     $N_a\times$16.06       & Ours                 &          \textbf{0.46}       &     \textbf{6.55}   \\ \hline
\end{tabular}
\caption{Computation Complexity $(\times{10^6})$ at first layer: $N_a$ is the number of events input to the feature extraction layer of \cite{1g}. Note the huge reduction in computation in EventF2S compared to other methods.}
\label{table:compute}
\end{table}


\section{Conclusion}

In this paper, we proposed an asynchronous, sparse-spiking, neuromorphic compatible AER sensor object recognition framework known as EventF2S. The pipeline of EventF2S comprises an event encoding layer followed by an AER recognition network. The event encoding layer of EventF2S (SS-TE) was specifically derived to convert raw and noisy AER sensor data into very few informative early spikes. Experiments conducted demonstrated the denoising capability of the SS-TE layer and how it contributed to sparse spiking and resultant reduced computation. Sparse spiking can lead to information loss, which has been compensated by utilizing learnable feature extraction spiking neurons. Experiments on neuromorphic datasets revealed the above-mentioned fact by exhibiting competitive recognition accuracy at par with state-of-the-art temporal encoded SNNs. Note being neuromorphic hardware friendly, EventF2S exploits the unique advantages of AER sensors. To summarize, EventF2S provides an object recognition solution aligned with the benefits of AER sensors, thus rendering itself more suitable for resource-constrained applications.


\newpage

\begin{thebibliography}{34}








\bibitem{8} Guo, Shasha, and Tobi Delbruck. Low cost and latency event camera background activity denoising. IEEE Transactions on Pattern Analysis and Machine Intelligence.

\bibitem{9} Y. Hu, S.-C. Liu, and T. Delbruck, “v2e: From video frames to realistic DVS events,” in 2021 IEEE/CVF Conference on Computer Vision and Pattern Recognition Workshops (CVPRW), IEEE, 2021

\bibitem{10} Feng Y and et al. Event density based denoising method for dynamic vision sensor. Applied Sciences,

\bibitem{11} Liu H and et al. Design of a spatiotemporal correlation filter for event-based sensor. IEEE International Symposium on Circuits and Systems, pages 722–725, 2015

\bibitem{12} M. Mahowald, “VLSI analogs of neuronal visual processing: A synthesis of form and function,” Ph.D. dissertation, California Institute of Technology, Pasadena, CA, May 1992

\bibitem{13} G. Gallego, T. Delbruck, G. M. Orchard, et al., “Event-based vision: A survey,” IEEE Trans. Pattern Anal. Mach. Intell., pp. 1–1, 2020

\bibitem{14} P. Lichtsteiner, C. Posch, and T. Delbruck, “A 128x128 120dB 15us latency asynchronous temporal contrast vision sensor,” IEEE J. Solid-State Circuits, vol. 43, no. 2, pp. 566–576, Feb. 2008

\bibitem{15} Wei Fang, Zhaofei Yu, Yanqi Chen, Timothée Masquelier, Tiejun Huang, and Yonghong Tian. In- corporating learnable membrane time constant to enhance learning of spiking neural networks. In Proceedings of the IEEE/CVF International Conference on Computer Vision, pp. 2661–2671, 2021

\bibitem{16} Yujie Wu, Lei Deng, Guoqi Li, Jun Zhu, Yuan Xie, and Luping Shi. Direct training for spiking neural networks: Faster, larger, better. In Proceedings of the AAAI Conference on Artificial Intelligence, volume 33, pp. 1311–1318, 2019

\bibitem{17} Ali Samadzadeh, Fatemeh Sadat Tabatabaei Far, Ali Javadi, Ahmad Nickabadi, and Morteza Haghir Chehreghani. Convolutional spiking neural networks for spatio-temporal feature extraction. arXiv preprint arXiv:2003.12346, 2020

\bibitem{18} Chankyu Lee, Adarsh Kumar Kosta, Alex Zihao Zhu, Kenneth Chaney, Kostas Daniilidis, and Kaushik Roy. Spike-flownet: event-based optical flow estimation with energy-efficient hybrid neural networks. In European Conference on Computer Vision, pp. 366–382. Springer, 2020

\bibitem{19} Nitin Rathi, Gopalakrishnan Srinivasan, Priyadarshini Panda, and Kaushik Roy. Enabling deep spiking neural networks with hybrid conversion and spike timing dependent backpropagation. arXiv preprint arXiv:2005.01807, 2020

\bibitem{20} Guillaume Bellec, Darjan Salaj, Anand Subramoney, Robert Legenstein, and Wolfgang Maass. Long short-term memory and learning-to-learn in networks of spiking neurons. Advances in neural information processing systems, 31, 2018

\bibitem{21} Yujie Wu, Lei Deng, Guoqi Li, Jun Zhu, and Luping Shi. Spatio-temporal backpropagation for training high-performance spiking neural networks. Frontiers in neuroscience, 12:331, 2018.

\bibitem{22} Dongsung Huh and Terrence J Sejnowski. Gradient descent for spiking neural networks. Advances in neural information processing systems, 31, 2018.


Systems.


\bibitem{25} P. Lichtsteiner, C. Posch, and T. Delbruck, A 128×128 120 dB 15$\mu$s latency asynchronous temporal contrast vision sensor, IEEE J. Solid-State Circuits, vol. 43, no. 2, pp. 566–576, 2008

\bibitem{26} C. Posch, D. Matolin, and R. Wohlgenannt, A QVGA 143 dB dynamic range frame-free PWM image sensor with lossless pixel-level video compression and time-domain CDS, IEEE J. Solid-State Circuits, vol. 46, no. 1, pp. 259–275, Jan. 2011

\bibitem{27} P. Lichtsteiner and T. Delbruck, “64x64 event-driven logarithmic temporal derivative silicon retina,” in IEEE Workshop on Charge-Coupled Devices and Advanced Image Sensors, 2005, pp. 157–160.

\bibitem{28} P. Lichtsteiner, “An AER temporal contrast vision sensor,” Ph. D. Thesis, ETH Zurich, Dept. of Physics (D-PHYS), Zurich, Switzerland, 2006.

\bibitem{29} M. Cook, L. Gugelmann, F. Jug, C. Krautz, and A. Steger, “Interacting maps for fast visual interpretation,” in Int. Joint Conf. Neural Netw. (IJCNN), 2011, pp. 770–776

\bibitem{30} H. Rebecq, T. Horstschafer, G. Gallego, and D. Scaramuzza, EVO: A geometric approach to event-based 6-DOF parallel tracking and mapping in real-time, IEEE Robot. Autom. Lett., vol. 2, no. 2, pp. 593–600, 2017.

\bibitem{31} M. Liu and T. Delbruck, “Adaptive time-slice block-matching optical flow algorithm for dynamic vision sensors,” in British Mach. Vis. Conf. (BMVC), 2018.

\bibitem{32} J. Kogler, C. Sulzbachner, and W. Kubinger, “Bio-inspired stereo vision system with silicon retina imagers,” in Int. Conf. Comput. Vis. Syst. (ICVS), 2009, pp. 174–183.

\bibitem{33} A. I. Maqueda, A. Loquercio, G. Gallego, N. Garcıa, and D. Scaramuzza, “Event-based vision meets deep learning on steering prediction for self-driving cars,” in IEEE Conf. Comput. Vis. Pattern Recog. (CVPR), 2018



\bibitem{36} Hesham Mostafa, Supervised learning based on temporal coding in spiking neural networks, IEEE Transactions on Neural Network and Learning Systems, 2017.

\bibitem{37} Lagorce, Xavier, et al. "Hots: a hierarchy of event-based time-surfaces for pattern recognition." IEEE transactions on pattern analysis and machine intelligence 39.7 (2016): 1346-1359.

\bibitem{38} Sironi, Amos, et al. "HATS: Histograms of averaged time surfaces for robust event-based object classification." Proceedings of the IEEE conference on computer vision and pattern recognition. 2018.

\bibitem{39} Serrano-Gotarredona, Teresa, and Bernabé Linares Barranco. "Poker-DVS and MNIST-DVS. Their history, how they were made, and other details." Frontiers in neuroscience 9, 481, 2015.

\bibitem{1a} T. Serre, L. Wolf, S. Bileschi, M. Riesenhuber, and T. Poggio, “Robust object recognition with cortex-like mechanisms,” Pattern Analysis and Machine Intelligence, IEEE Transactions on, vol. 29, no. 3, pp. 411–426, 2007

\bibitem{1b} S. Chen, P. Akselrod, B. Zhao, J. A. P. Carrasco, B. Linares-Barranco, and E. Culurciello, “Efficient feedforward categorization of objects and human postures with address-event image sensors,” IEEE Transactions on Pattern Analysis and Machine Intelligence, vol. 34, no. 2, pp. 302–314, 2012.

\bibitem{1c} B. Zhao, R. Ding, S. Chen, B. Linares-Barranco, and H. Tang, “Feed-forward categorization on aer motion events using cortex-like features in a spiking neural network,” IEEE transactions on neural networks and learning systems, vol. 26, no. 9, pp. 1963–1978, 2014

\bibitem{1d} Garrick Orchard, Cedric Meyer, Ralph Etienne-Cummings, Christoph Posch,
Nitish Thakor, Ryad Benosman, "Hfirst: A temporal approach to object recognition,” IEEE Transactions on Pattern Analysis and Machine Intelligence, 2015.

\bibitem{1d-1} G. Orchard, A. Jayawant, G. K. Cohen, and N. Thakor, “Converting static image datasets to spiking neuromorphic datasets using saccades,” Frontiers in Neuroscience, vol. 9, p. 437, 2015.

\bibitem{1e} Cohen, G. K, Orchard, G, Leng, S. H, Tapson, J, Benosman, R. B and Van Schaik, A. 2016. Skimming digits: neuromorphic classiﬁcation of spike-encoded images. Fron- tiers in neuroscience 10:184.

\bibitem{1f} Liu, Q, Pan, G, Ruan, H, Xing, D, Xu, Q, and Tang, H, Unsupervised aer object recognition based on multiscale spatio-temporal features and spiking neurons. arXiv preprint arXiv:1911.08261, 2019

\bibitem{1g} R. Xiao, H. Tang, Y. Ma, R. Yan, and G. Orchard, An event-driven categorization model for aer image sensors using multispike encoding and learning, IEEE Transactions on Neural Networks and Learning
Systems, 2020

\bibitem{1h} Qianhui Liu, Haibo Ruan, Dong Xing, Huajin Tang, Gang Pan, Effective AER Object Classification Using Segmented Probability-Maximization Learning in Spiking Neural Networks, AAAI, 2020

\bibitem{2a} X. Peng, B. Zhao, R. Yan, H. Tang, and Z. Yi, “Bag of events: An efficient probability-based feature extraction method for aer image sensors,” IEEE Transactions on Neural Networks and Learning Systems, vol. 28, no. 4, pp. 791–803, 2017

\bibitem{2b} Lagorce, X.; Orchard, G.; Galluppi, F.; Shi, B. E.; and Benosman, R. B. 2017. Hots: a hierarchy of event-based time-surfaces for pattern recognition. IEEE Transactions on Pattern Analysis and Machine Intelligence 39(7):1346– 1359.

\bibitem{3a} Shibo Zhou, Wei Wang, Xiaohua Li and Zhanpeng Jin, "A spike learning system for event driven object recognition", 2021

\end{thebibliography}
\end{document}